\documentclass[conference]{IEEEtran}
\IEEEoverridecommandlockouts

\usepackage{epsfig}
\usepackage{subcaption}
\usepackage{calc}
\usepackage{amssymb}
\usepackage{amstext}
\usepackage{amsmath}
\usepackage{amsthm}
\usepackage{multicol}
\usepackage{pslatex}
\usepackage{algorithm2e}
\usepackage[bottom]{footmisc}
\usepackage{makecell}
\usepackage{tikz}
\usepackage{xtab,afterpage}
\usepackage{longtable}
\usetikzlibrary{shapes.geometric, arrows, calc}

\ifCLASSINFOpdf
\else
\fi

\begin{document}
\title{Ontology engineering with Large Language Models}

\author{
    \IEEEauthorblockN{Patricia Mateiu\IEEEauthorrefmark{1} and Adrian Groza\IEEEauthorrefmark{1}}
    \IEEEauthorblockA{\IEEEauthorrefmark{1}Department of Computer Science, \\
    Technical University of Cluj-Napoca, 400114 Cluj-Napoca, Romania
    \\\{patriciamateiu33@gmail.com, adrian.groza@cs.utcluj.ro\}}
    
}

\maketitle

\begin{abstract}
We tackle the task of enriching ontologies by automatically translating natural language sentences into Description Logic. 
Since Large Language Models (LLMs) are the best tools for translations, we fine-tuned a GPT-3 model to convert Natural Language sentences into OWL Functional Syntax. 
We employ objective and concise examples to fine-tune the model regarding: instances, class subsumption, domain and range of relations, object properties relationships, disjoint classes, complements, cardinality restrictions.
The resulted axioms are used to enrich an ontology, in a human supervised manner. 
The developed tool is publicly provided as a Prot\'{e}g\'{e}\ plugin.
\end{abstract}

\IEEEpeerreviewmaketitle

\section{Motivation}
\label{sec:introduction}
The technical challenges and costs associated with the development of ontologies are arguable the main causes for the partial failure of the Semantic Web. 
Aiming to facilitate the development of ontologies by industry, the economical aspects of ontology engineering have been subject to the Ontology Cost Model (ONTOCOM)~\cite{paslaru2006ontocom}. 
Despite the existence of several ontology engineering methodologies (OEMs) (no less than 15 as identified by~\cite{iqbal2013analysis} in 2013), the domain of Semantic Web does not benefit from a mature and largely accepted methodology. 
More recently,~\cite{kotis2020ontology} have analysed 9 OEMs, concluding that non-collaborative methodologies have a negative impact on the liveness, evolution, and reusability of the ontologies.

We rely here on the current opportunities provided by Large Language Models (LLMs). 
We argue that LLMs have the potential to largely increase the efficiency of the ontology engineering. 
Since LLMs are best technology for language translation, our employ them to translate from natural language to description logic (DL). 
That is, given a description in natural language (definition, domain knowledge), the aim is to automatically obtain corresponding ontology in a formal language. 
We developed a tool able to enrich and populate an ontology with domain knowledge available in natural language. 
The tool relies on GPT-3 which we fine-tuned for the ontology engineering task. 
The solution is freely available and is provided as a plugin for the Prot\'{e}g\'{e}\ editor.

\section{Tuning the model for OWL}
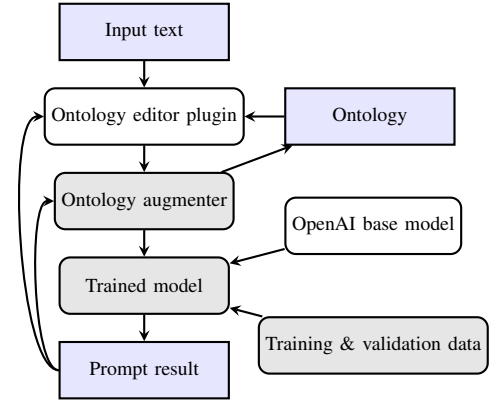
\begin{figure}
    \centering
    \tikzstyle{io} = [rectangle, minimum width=3cm, minimum height=1cm, text centered, draw=black, fill=blue!10]
\tikzstyle{code} = [rectangle, rounded corners, minimum width=3cm, minimum height=1cm, text centered, draw=black, fill=gray!20]
\tikzstyle{visible} = [rectangle, rounded corners, minimum width=3cm, minimum height=1cm,text centered, draw=black, fill=white!10]
\tikzstyle{arrow} = [thick,->,>=stealth]
\tikzstyle{loop} = [thick,->,>=stealth]
\begin{tikzpicture}[thick,scale=0.75, every node/.style={scale=0.75},node distance=1.5cm]
\node (n1) [io] {Input text};
\node (n2) [visible, below of=n1] {Ontology editor plugin};
\node (n3) [code, below of=n2] {Ontology augmenter};
\node (n4) [code, below of=n3] {Trained model};
\node (n5) [visible, above right of=n4, xshift=3cm] {OpenAI base model};
\node (n6) [code, below right of=n4, xshift=3cm] {Training \& validation data};
\node (n7) [io, below of=n4] {Prompt result};
\node (n8) [io, right of=n2, xshift=2.5cm] {Ontology};
\draw [arrow] (n1) -- (n2);
\draw [arrow] (n2) -- (n3);
\draw [arrow] (n3) -- (n4);
\draw [arrow] (n5) -- (n4);
\draw [arrow] (n6) -- (n4);
\draw [arrow] (n4) -- (n7);
\draw [arrow] (n8) -- (n2);
\draw [arrow] (n3) -- (n8);
\draw [loop]  (n7) to [out=-180,in=-180,loop,looseness=0.4] node[anchor=north]{}(n2);
\draw [loop]  (n7) to [out=-180,in=-180,loop,looseness=0.4] node[anchor=north]{}(n3);
\end{tikzpicture}
\caption{Data flow for the presented approach}
    \label{fig:flow}
\end{figure}

The tool is constructed as a Prot\'{e}g\'{e} plugin that supports the development of an ontology from scratch and also the enrichment of an existing ontology. 
Natural language sentences are translated into ontological elements and appended to the current ontology (Figure \ref{fig:flow}). 
The prompts are sent to a fine-tuned GPT-3 davinci model which returns the result into ontology axioms. 
Technically, these axioms are handled using the \textit{owlapi} Java library~\cite{horridge2011owl}, and appended to the active ontology in the Prot\'{e}g\'{e} editor.

We developed a dataset of 150 pairs of prompts and their corresponding translations into OWL Functional Syntax. 
We seek to cover various cases and to incorporate a variety of domains, in favor of attaining an adaptable model. 
Tables~\ref{tab:example1} and \ref{tab:example2} depict some of the prompts used in the training set. 
We used the following conventions:

First, the underscore symbol is used for elements having multiple words, since it is the default naming convention used in Prot\'{e}g\'{e}, e.g., the object property named \textit{has\_sibling} or compound individual names such as \textit{Wolfgang\_Amadeus\_Mozart}. 

Second, we settled the class and object property names in only lowercase letters, while the individual names can begin either with a lowercase or an uppercase letter.

Third, for subclass relationships, to highlight the general aspect, several connection words were used. For example, the sentence \textit{every man is a person} creates a subclass relationship between the class \textit{person} and the class \textit{man}. 
If these classes do not exist in the ontology, the declaration axioms will generate them. 
Otherwise, these axioms are ignored. 
The quantifier \textit{every} is used to emphasize that there is no instance where an individual that belongs to the class \textit{man} will not belong to the class \textit{person}. 
Other connection words can be used for defining this type of relationship, e.g.,  \textit{all lilies are flowers}, \textit{cats are a type of animal}. 
Multiple subclass relationships can be used at the same time, as in the example \textit{bugs, ants, ladybugs, flies are insects}.

\begin{table}
\caption{Fine-tuning davinci model for classes}
\label{tab:example1}
\centering
\scalebox{0.85}{
\begin{tabular}{|l|l|}
  \hline
  Prompt & Result \\
  \hline
  Anna is a girl & \makecell{Declaration(Class(:girl))\\Declaration(NamedIndividual(:Anna))\\ClassAssertion(:girl :Anna)}\\
  \hline
  every rose is a flower & \makecell{Declaration(Class(:flower))\\Declaration(Class(:rose))\\SubClassOf(:rose :flower)}\\
  \hline
  Brandon is a male artist & \makecell{Declaration(Class(:male))\\Declaration(Class(:artist))\\Declaration(NamedIndividual(:Brandon))\\ClassAssertion(:male :Brandon)\\ClassAssertion(:artist :Brandon)}\\
  \hline
  no woman is a man & \makecell{Declaration(Class(:man))\\Declaration(Class(:woman))\\DisjointClasses(:man :woman)}\\
  \hline
  Tina is not a cat & \makecell{Declaration(Class(:cat))\\Declaration(NamedIndividual(:Tina))\\ClassAssertion(ObjectComplementOf(:cat) :Tina)}\\
  \hline
  \end{tabular}
  }
  \end{table}

  \begin{table}
  \caption{Fine-tuning davinci model for object properties}
  \label{tab:example2}
  \centering
 \scalebox{0.82}{
  \begin{tabular}{|p{2.5cm}|l|}
  \hline
  Prompt & Result \\
  \hline
  Sarah and Anna are each other's sisters & \makecell{Declaration(ObjectProperty(:has\_sister))\\Declaration(NamedIndividual(:Anna))\\Declaration(NamedIndividual(:Sarah))\\ObjectPropertyAssertion(:has\_sister :Anna :Sarah)\\ObjectPropertyAssertion(:has\_sister :Sarah :Anna)}\\
  \hline
  Mia owns 2 bikes & \makecell{Declaration(Class(:bike))\\Declaration(ObjectProperty(:owns))\\Declaration(NamedIndividual(:Mia))\\ClassAssertion(ObjectExactCardinality(2 :owns :bike) :Mia)}\\
  \hline
  Michael owns tractors & \makecell{Declaration(Class(:tractor))\\Declaration(ObjectProperty(:owns))\\Declaration(NamedIndividual(:Michael))\\ClassAssertion(ObjectSomeValuesFrom(:owns :tractor) :Michael)}\\
  \hline
  rabbits eat only carrots & \makecell{Declaration(Class(:carrot))\\Declaration(Class(:rabbit))\\Declaration(ObjectProperty(:eats))\\SubClassOf(:rabbit ObjectAllValuesFrom(:eats :carrot))\\}\\
  \hline
  all animals have exactly one tail & \makecell{Declaration(Class(:animal))\\Declaration(Class(:tail))\\Declaration(ObjectProperty(:has))\\EquivalentClasses(:animal ObjectExactCardinality(1 :has :tail))}\\
  \hline
  a mother is a female who has at least 1 child & \makecell{Declaration(Class(:child))\\Declaration(Class(:female))\\Declaration(Class(:mother))\\Declaration(ObjectProperty(:has))\\EquivalentClasses(:mother ObjectIntersectionOf\\(:female ObjectMinCardinality(1 :has :child)))}\\
  \hline
  Penn's mother is Sonia & \makecell{Declaration(ObjectProperty(:has\_mother))\\Declaration(NamedIndividual(:Penn))\\Declaration(NamedIndividual(:Sonia))\\ObjectPropertyAssertion(:has\_mother :Penn :Sonia)}\\
  \hline
  Anna does not know Sabina & \makecell{Declaration(ObjectProperty(:knows))\\Declaration(NamedIndividual(:Anna))\\Declaration(NamedIndividual(:Sabina))\\NegativeObjectPropertyAssertion(:knows :Anna :Sabina)}\\
  \hline
  Jenna is a fan of Britney Spears & \makecell{Declaration(Class(:movie))\\Declaration(ObjectProperty(:is\_a\_fan\_of))\\Declaration(NamedIndividual(:Britney\_Spears))\\Declaration(NamedIndividual(:Jenna))\\ObjectPropertyAssertion(:is\_a\_fan\_of :Jenna :Britney\_Spears)}\\
  \hline
  Cora and Meena hate each other & \makecell{Declaration(ObjectProperty(:hates))\\Declaration(NamedIndividual(:Cora))\\Declaration(NamedIndividual(:Meena))\\ObjectPropertyAssertion(:hates :Cora :Meena)\\ObjectPropertyAssertion(:hates :Meena :Cora)}\\
  \hline
  anyone who is a sister is female & \makecell{Declaration(Class(:female))\\Declaration(ObjectProperty(:has\_sister))\\ObjectPropertyRange(:has\_sister :female)}\\
  \hline
  anybody who has a brother has a sibling & \makecell{Declaration(ObjectProperty(:has\_brother))\\Declaration(ObjectProperty(:has\_sibling))\\SubObjectPropertyOf(:has\_brother :has\_sibling)}\\
  \hline
  if X has friend Y then Y has friend X & \makecell{Declaration(ObjectProperty(:has\_friend))\\SymmetricObjectProperty(:has\_friend)}\\
  \hline
  if X has mother Y then Y does not have mother X & \makecell{Declaration(ObjectProperty(:has\_mother))\\AsymmetricObjectProperty(:has\_mother)}\\
  \hline
 \end{tabular}
 }
\end{table}

\section{Experiments}
\subsection{Running Scenario}
\begin{figure}
    \centering
    \includegraphics[width=0.47\textwidth]{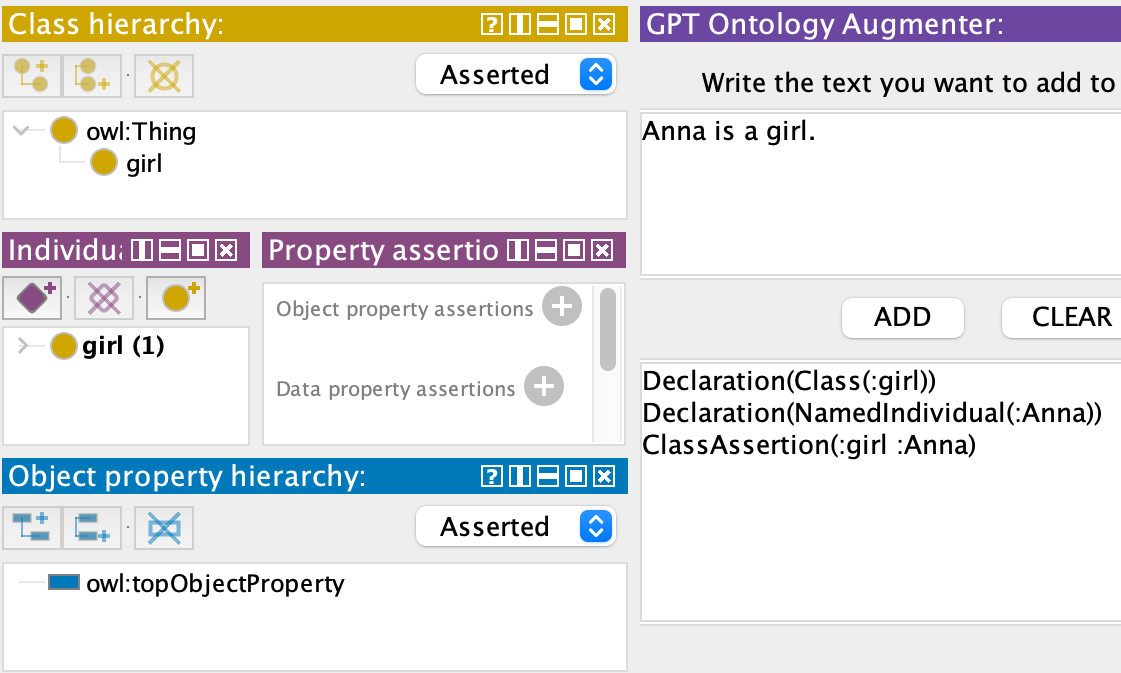}
    \caption{Adding new instances and classes}
    \label{fig:family_1}
\end{figure}

We exemplify the functionalities of the plugin on the family ontology. 
First, let the sentence \textit{$s_1$: Ana is a girl}, which the tool automatically translates in OWL Functional Syntax with three axioms (line 1 in Table~\ref{tab:example1}).
\begin{eqnarray}
    Declaration(Class(:girl))\\
    Declaration(NamedIndividual(:Anna))\\
    ClassAssertion(:girl :Anna)
\end{eqnarray}
Since the first axiom is a declaration, the new class, \textit{girl} is added to the taxonomy, as subclass of the top concept, \textit{owl:Thing} (see Figure~\ref{fig:family_1}).
The second axiom is also a declaration, but of an individual, and will be added to the \textit{Individuals by type} panel. 
Additionally, the class will appear in this panel as well, since the third axiom is a class assertion axiom, meaning that the individual \textit{Anna} is included in the class \textit{girl}. 
These axioms are added to a temporary ontology, and retrieved using \textit{OWLAPI} library.

\begin{figure}
    \centering
    \includegraphics[width=0.47\textwidth]{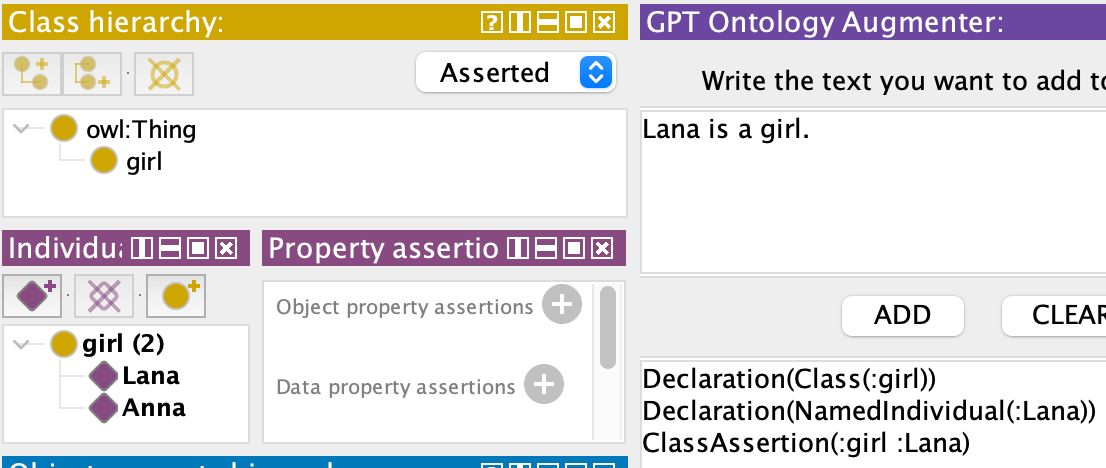}
    \caption{Adding individuals to existing classes}
    \label{fig:family_2}
\end{figure}

Second, let the sentence \textit{$s_2$: Lana is a girl}. 
This statement involves the same class as in $s_1$, but a different individual in the assertion. 
The ontology will be populated with the new individual \textit{Lana}, which belongs to the already existing class \textit{girl}. 
Figure \ref{fig:family_2} shows that the new individual is added to the list corresponding to class defined in the first step.

\begin{figure}
    \centering
    \includegraphics[width=0.47\textwidth]{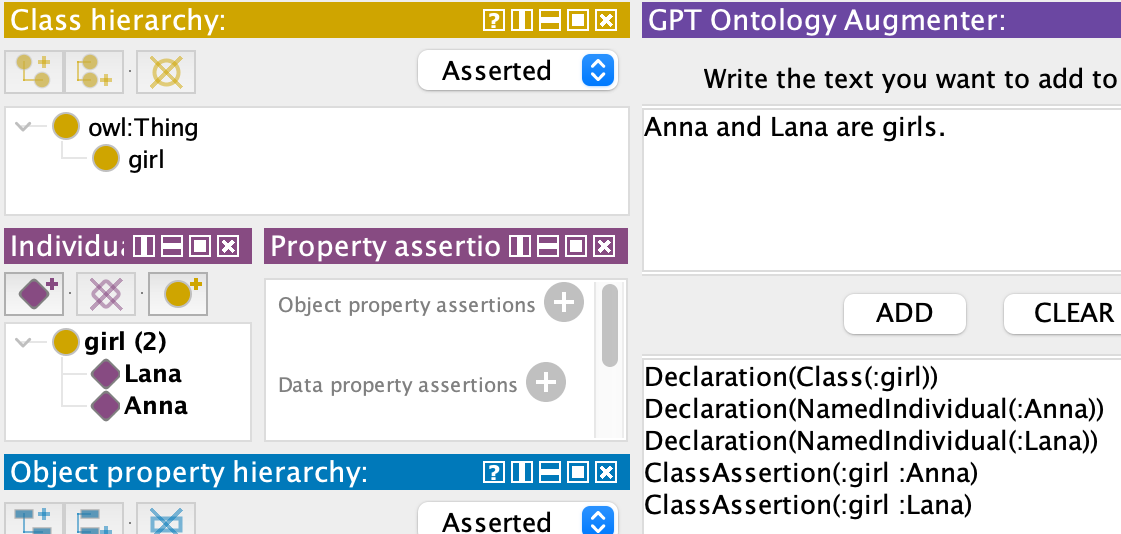}
    \caption{Formalising multiple individuals belonging to the same collective class }
    \label{fig:family_3}
\end{figure}

Third, let the alternative sentence \textit{$s_3$: Anna and Lana are girls}. 
Here, instead of specifying the class for each named individual, multiple individuals belong to the same collective class \textit{girl}.
The plugin is able to deliver the same ontology as in case of statements $s_1$ and $s_2$ (see Figure~\ref{fig:family_3}).  
The class is declared only once, while the second declaration of the same class is ignored. 
The trained model offers the users several options in transmitting the components they want to include in the ontology. 

\begin{figure}
    \centering
    \includegraphics[width=0.47\textwidth]{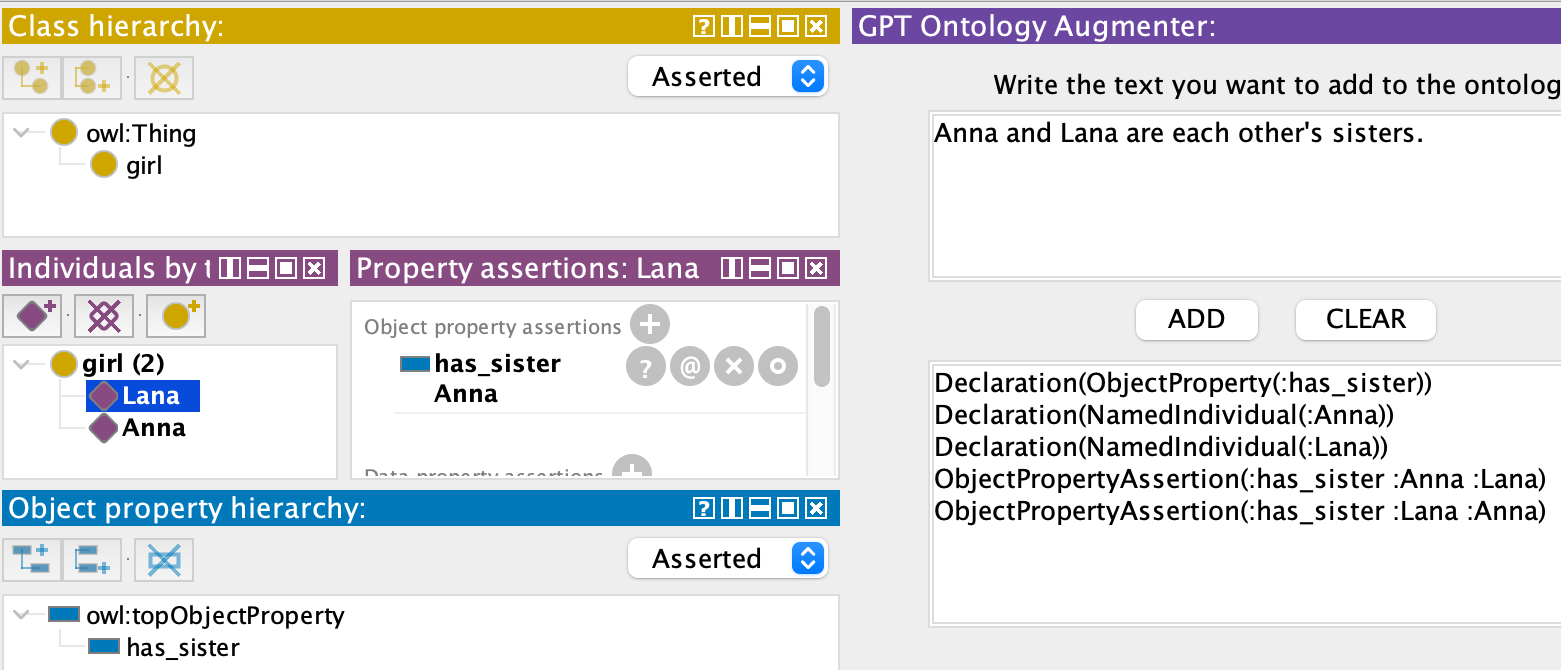}
    \caption{Handling symmetric relations}
    \label{fig:family_4}
\end{figure}

Fourth, one can add object properties. Consider the statement $s_4$: \textit{Anna and Lana are each other's sisters}. 
The resulted property is attached to the object property taxonomy, and the assertions can be seen in the third panel of the left half in Figure~\ref{fig:family_4}, by clicking on each individual. 
Figure \ref{fig:family_4} presents the relation \textit{Lana has sister Anna} and the inverse relation. 

\begin{figure}
    \centering
    \includegraphics[width=0.47\textwidth]{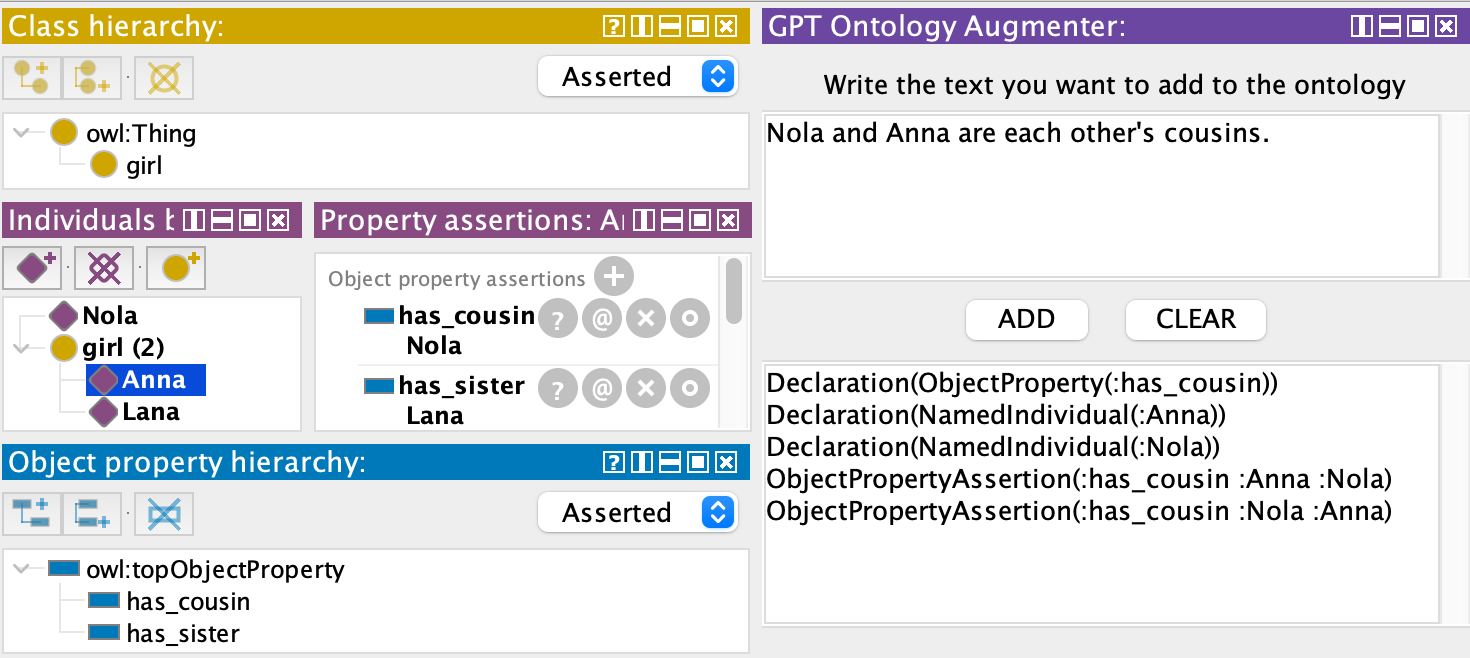}
    \caption{Adding assertions about new individuals}
    \label{fig:family_6}
\end{figure}

Fifth, one can add assertions about new individuals. 
Let $s_5$: \textit{Nola and Anna are each other's cousins}. 
In this case, \textit{Nola}, who was not defined prior and has no class association, will be added as an individual, but separate than the ones grouped by class. 
The object property will be added for both individuals, just like in the previous step (see Figure~\ref{fig:family_6}). 

\subsection{Prompt engineering strategies}
We run experiments with three strategies: zero-shot  learning, few-shot learning and fine-tuning.

The \textit{Zero-shot learning} strategy asks the language model to generate the output directly, with no presented examples. 
Several experiments were run with this strategy, using the GPT-3.5-turbo model. 
The results were, although not incorrect, not the expected ones. 
For example, using prompt \textit{Translate 'Anna is a girl' into Functional Syntax}, the model returned \textit{is(Anna, girl)}, which is not helpful since it is not in the form of an axiom. 
It rather represents the relationship \textit{is} between \textit{Anna} and \textit{girl}, yet does not offer any information on the form or meaning of these words.

The \textit{few-shot learning} strategy lets LLMs train for specific tasks from a few examples.
To assess this strategy, we tested various prompts. 
The first trial was \textit{Anna is a girl}, whose corresponding axioms are also included in the example. 
The outcome for this prompt was not the expected one and the solution was proven to be inconsistent, returning different results for the same prompt. In the first trial, besides the declarations and the class assertion for individual \textit{Anna} and class \textit{girl}, the response includes other axioms, that are not logical for the given context. Namely, the last line is an object property assertion between an individual and a class, which is not possible. The second trial is not as faulty, it does not include incorrect axioms, but it includes extra axioms that are not needed. For example, the declaration for the class \textit{person} and the object property \textit{married\_to}, which are not in the prompt. Results such as this one would cause users to have an ontology that is too big, and only partially used, which, in fact, contradicts their preferences and intentions.

The \textit{fine-tuning} strategy requires a dedicated dataset. 
A dataset with 150 prompt-result pairs and a validation set with 50 such pairs were used, with the same format and variation as those in the training data set, but with different cases and examples. 
The validation data set is used to determine the optimal combination of hyper-parameters that would have the best token accuracy. Regarding the data structure, several tests were done to determine the correct order of the completion result with respect to the prompt. One question was whether it was better to write the declarations and assertion or relationship axioms in the order that the words appear in the sentence or to write them in the order that Prot\'{e}g\'{e} would save them in a Functional Syntax. 
Both options were considered in training with several combinations of hyper-parameters. 

\section{Related Work} \label{related-work}
Before the LLMs era tools like Fred~\cite{draicchio2013fred} were used. 
Fred is a machine reader designed for the Semantic Web that can analyze natural language in 48 different languages and generates linked data in the OWL format. 
However, the resulting axioms require further processing before they can be effectively utilized for reasoning.
In the recent years, there have been various approaches in learning ontologies from text data, by extracting the ontological terms and structuring them into one component~\cite{al2020automatic}. 
For instance,  Bo\v{z}i{\'c} \cite{bovzicsemantic} has analysed the potential combining Semantic Web and GPT, as well as the related risks that might pose a threat. 

OntoGPT tool~\cite{caufield2023structured} extracts information from text, by using three strategies: SPIRES, HALO, and SPINDOCTOR.
SPIRES applies a knowledge schema on the input text and returns an instance with multiple attribute-value relations, where the values are either data primitives or other instances, thus creating a linked scheme.
HALO is a Few-Shot Learning approach, which solves tasks with limited number of examples for learning while using prior knowledge.

Conceptual modelling using large language models have been experienced by~\cite{fill2023conceptual}.  
ChatGPT was used to generate entity-relations diagrams (ER). 
The designed prompt starts with an explanation of ER diagrams. 
Then the prompt includes an example of ER diagram in JSON syntax. 
The last part of the prompt is the natural language description of the task. 
A second experiment has focused on business process diagrams. 
A subset of BPMN diagrams has been considered. 
The prompt describes the meta-model in NL (e.g. a task has exactly one predecessor and one successor) and an example in JSON format.  
The third experiment has targeted UML class diagrams, for which a Zero-Shot approach has been preferred.
Even if large parts of the conceptual modelling was correct, modelling experience of a human expert was required to validate the model. 

GraphGPT~\cite{graphgpt} converts natural language into knowledge graphs. The application does not imply ontology population, it rather offers the users a view on how the data they submit might be connected. 
Bikeyev \cite{bikeyev2023synthetic} has proposed an alternative of knowledge model engineering and knowledge graph generation as an automated approach that avoids the vagueness of Natural Language Processing. A bottom-up approach is combined to a LLM, namely GPT-3. 
The method uses two types of prompts, one to generate a hierarchy of elements and the other to determine possible relationships between them. In both cases, it is necessary that the prompts respect memory limitations, so that the prompt and the result can fit entirely in the given memory slot. After the initial hierarchy is constructed, each element can be used in another prompt to give a more detailed result, and this step can continue until the result is satisfactory. 
The advantage is that this approach can suit the individual preferences of each user, depending on how much detail they need in the ontology. 
GraphGPT can incorporate newly available data when updating, thus allowing a form of stream reasoning~\cite{groza2012plausible} when populating the knowledge graph. 

Csaszar and Slavescu have developed a tool to help software developers visualize the call graph of their code while editing it. 
Two graphs are automatically built from the source code: the import graph and the call graph. 
Instead of LLMs, the process of building the two graphs is based on using a query based architecture, commonly used by language servers
~\cite{csaszar2020interactive}

In a literature dominated by transformers, Ilies and Marginean have used grammars to provide a white box alternative~\cite{ilies2021understanding}. 
Context free grammars and semantic roles are used to structure knowledge from texts related to cooking recipes. 
Lex and Yacc interleave with AllenNLP to compute a parse tree for a cooking recipe, where each group of words is labeled with an appropriate semantic role. 
The approach can be applied to other types of instruction manuals.

Yang et al. \cite{yang2023harnessing} have developed LOGICLLAMA, a fine-tuned tool used for natural language to First-Order Logic translation, which can be also used for correcting FOL results generated by GPT-3.5 and is comparable to GPT-4. The MALLS dataset contains pairs NL-FOL generated by GPT-4 and is intended to be used for fine-tuning and testing the model. These pairs are resulted by repeatedly prompting GPT-4 using a pipeline which adjusts depending on the previous results. The LLOGICLLAMA is obtained from training and fine-tuning a model using the MALLS data set for two main tasks, generating translation from NL to FOL and correcting already generated translations by GPT-3.5. 
The first one uses natural language text as input and provides FOL output, while the second one uses a pair of NL and the resulted FOL translation returned by GPT-3.5 and provides a single output in FOL, representing the necessary adjustments or corrections. 
 
\section{Conclusion}
The developed plugin shows hows language models (e.g. GPT) can be used in automating learning and populating ontologies, a process which is very time-consuming, complex and could be overwhelming in terms of decision making. 
The aim was to exploit the capabilities of pre-trained language models to obtain OWL axioms.
 Aware of the limitations and risks of Large Language Models, the tool aims to be a support tool that saves development time. 
 It also reduces the interaction time with the domain expert, given that a description of the domain exists in natural language. 
 Ongoing work regards quantitative evaluation and assessing the efficiency of ontology engineering with and without the tool. 
\bibliography{bib}
\bibliographystyle{IEEEtran}
\end{document}